\documentclass[a4paper,conference]{IEEEtran}
\IEEEoverridecommandlockouts
\usepackage{url}
\usepackage{graphicx}
\usepackage{subcaption}
\usepackage{booktabs}
\usepackage{amsmath}
\usepackage{amssymb}
\usepackage{mathtools}
\usepackage{multirow}
\usepackage{color}

\usepackage{blindtext}
\usepackage{multicol}

\ifCLASSINFOpdf
\else
\fi
\begin{document}
%
\newtheorem{example}{Example}[section]
\newtheorem{problem}{Problem}[section]
\title{Cross-Supervised Joint-Event-Extraction with Heterogeneous Information Networks}

\author{\IEEEauthorblockN{Yue Wang$^{1,5}$, Zhuo Xu$^1$, Lu Bai$^1$\IEEEauthorrefmark{1}\thanks{\IEEEauthorrefmark{1}Lu Bai is the corresponding author.}, Yao Wan$^2$, Lixin Cui$^1$, Qian Zhao$^5$, Edwin R. Hancock$^3$, Philip S. Yu$^4$}
\IEEEauthorblockA{$^{1}$ Central University of Finance and Economics, Beijing, China\\
$^{2}$Huazhong University of Science and Technology, Wuhan, China\\
$^{3}$Department of Computer Science, University of York, York, UK\\
$^{4}$University of Illinois at Chicago, Chicago, USA\\
$^{5}$State Key Laboratory of Cognitive Intelligence, iFLYTEK, Hefei, China\\
\{wangyuecs,bailucs,cuilixin\}@cufe.edu.cn, wanyao@hust.edu.cn,\\
 qianzhao@iflytek.com, edwin.hancock@york.ac.uk, psyu@uic.edu\\
}
}


%


\maketitle

\begin{abstract}
Joint-event-extraction, which extracts structural information (i.e., entities or triggers of events) from unstructured real-world corpora, has attracted more and more research attention in natural language processing. Most existing works do not fully address the sparse co-occurrence relationships between entities and triggers, which loses this important information and thus deteriorates the extraction performance. To mitigate this issue, we first define the joint-event-extraction as a sequence-to-sequence labeling task with a tag set composed of tags of triggers and entities. Then, to incorporate the missing information in the aforementioned co-occurrence relationships, we propose a \underline{C}ross-\underline{S}upervised \underline{M}echanism (CSM) to alternately supervise the extraction of either triggers or entities based on the type distribution of each other. Moreover, since the connected entities and triggers naturally form a heterogeneous information network (HIN), we leverage the latent pattern along meta-paths for a given corpus to further improve the performance of our proposed method. To verify the effectiveness of our proposed method, we conduct extensive experiments on four real-world datasets as well as compare our method with state-of-the-art methods. Empirical results and analysis show that our approach outperforms the state-of-the-art methods in both entity and trigger extraction.
\end{abstract}

\begin{IEEEkeywords}
Joint-event-extraction, neural networks, heterogeneous information network
\end{IEEEkeywords}

%
\IEEEpeerreviewmaketitle

\section{Introduction}
Event extraction~\cite{DBLP:conf/kdd/RitterMEC12} is a process to extract the named entities~\cite{DBLP:conf/naacl/LampleBSKD16}, event triggers~\cite{DBLP:conf/acl/BronsteinDLJF15} and their relationships from real-world corpora. The named entities refer to those texts about predefined classes (e.g. person names, company name and locations) and event triggers are words that express the types of events in texts~\cite{DBLP:conf/acl/BronsteinDLJF15} (e.g., the word ``hire'' may trigger an ``employ'' event type). In literature, named entities and triggers are connected and named entities with corresponding roles are called arguments for a given trigger~\cite{DBLP:conf/naacl/ChenJ09} of a specific event.


Currently, most existing works divide the event extraction into two independent sub-tasks: named entity recognition~\cite{DBLP:conf/naacl/LampleBSKD16} and trigger labeling~\cite{DBLP:conf/acl/BronsteinDLJF15}. These two sub-tasks are always formulated as multi-class classification problems, and many works apply the sequence-to-sequence based labeling method which aims to translate a sentence into sequential tags~\cite{DBLP:journals/tacl/ChiuN16}. From our investigation, one problem of these sequence-to-sequence methods is that they ignore the orders of output tags, and therefore, it is difficult to precisely annotate different parts of an entity. To address this issue, some methods~\cite{MaH16,DBLP:conf/www/AlzaidyCG19} propose to incorporate the conditional random field (CRF) module to be aware of order-constraints for the annotated tags.

Since entities and triggers are naturally connected around events, recent works try to extract them jointly from corpora. Early methods apply pipeline frameworks with predefined lexical features~\cite{DBLP:conf/acl/LiJH13} which lack generality to different applications. Recent works leverage the structural dependency between entities and triggers~\cite{DBLP:conf/naacl/YangM16,DBLP:conf/ijcai/ZhangQZLJ19} to further improve the performances of both the entity and trigger identification sub-tasks.


Although existing works have achieved comparable performance on jointly extracting entities and triggers, these approaches still suffer the major limitation of
losing co-occurrence relationships between entities and triggers. Many existing methods determine the trigger and entities separately and then match the entities with triggers~\cite{DBLP:conf/naacl/YangM16,DBLP:conf/aaai/NguyenN19}.
In this way, the co-occurrence relationships between entities and triggers are ignored, although pre-trained features or prior data are introduced to achieve better performance.
It is also challenging to capture effective co-occurrence relationships between the entities and their triggers. We observed from the experiments that most of the entities and triggers are co-occurred sparsely (or indirectly) throughout a corpus. This issue exacerbates the problem of losing co-occurrence relationships mentioned before.


\begin{figure*}[htb]
	\centering
\begin{subfigure}{2.8in}
		\includegraphics[width=\textwidth]{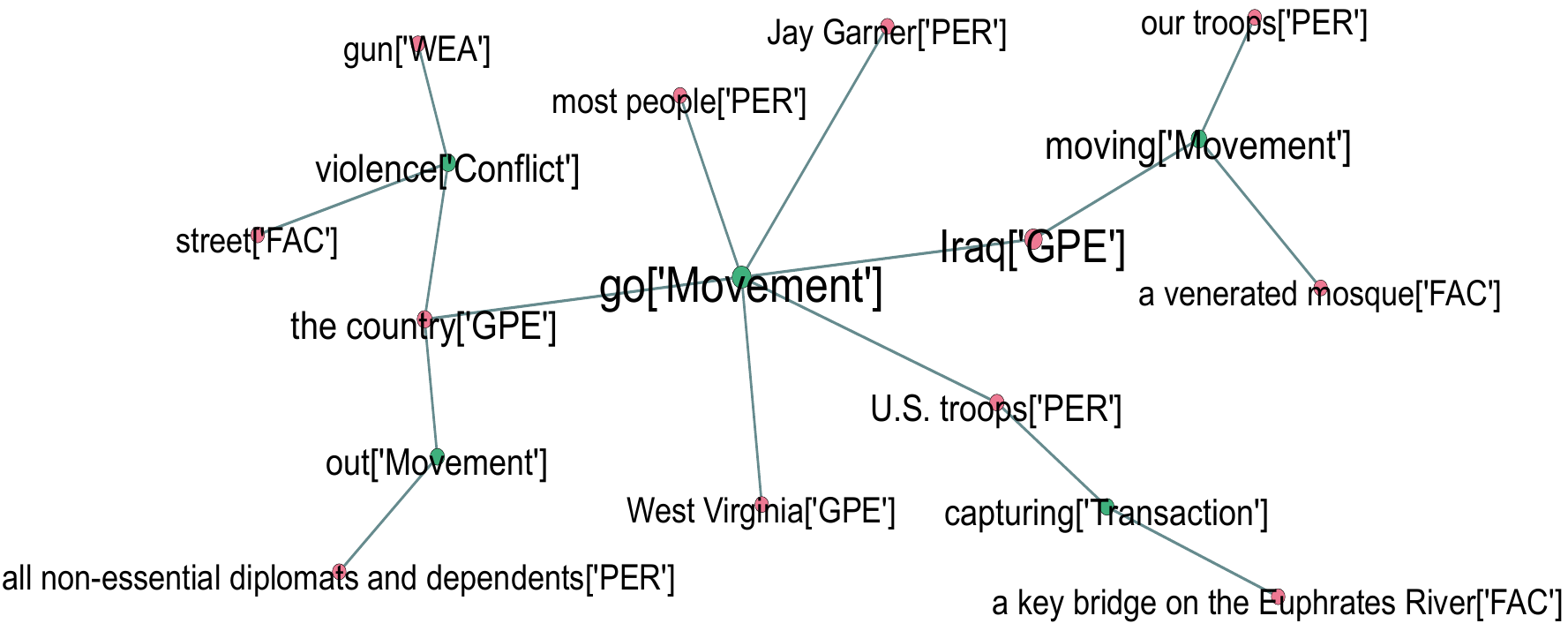}
	\caption{}
	\label{fig:illustrated_a} 
\end{subfigure}
\begin{subfigure}{1.8in}
		\includegraphics[width=\textwidth]{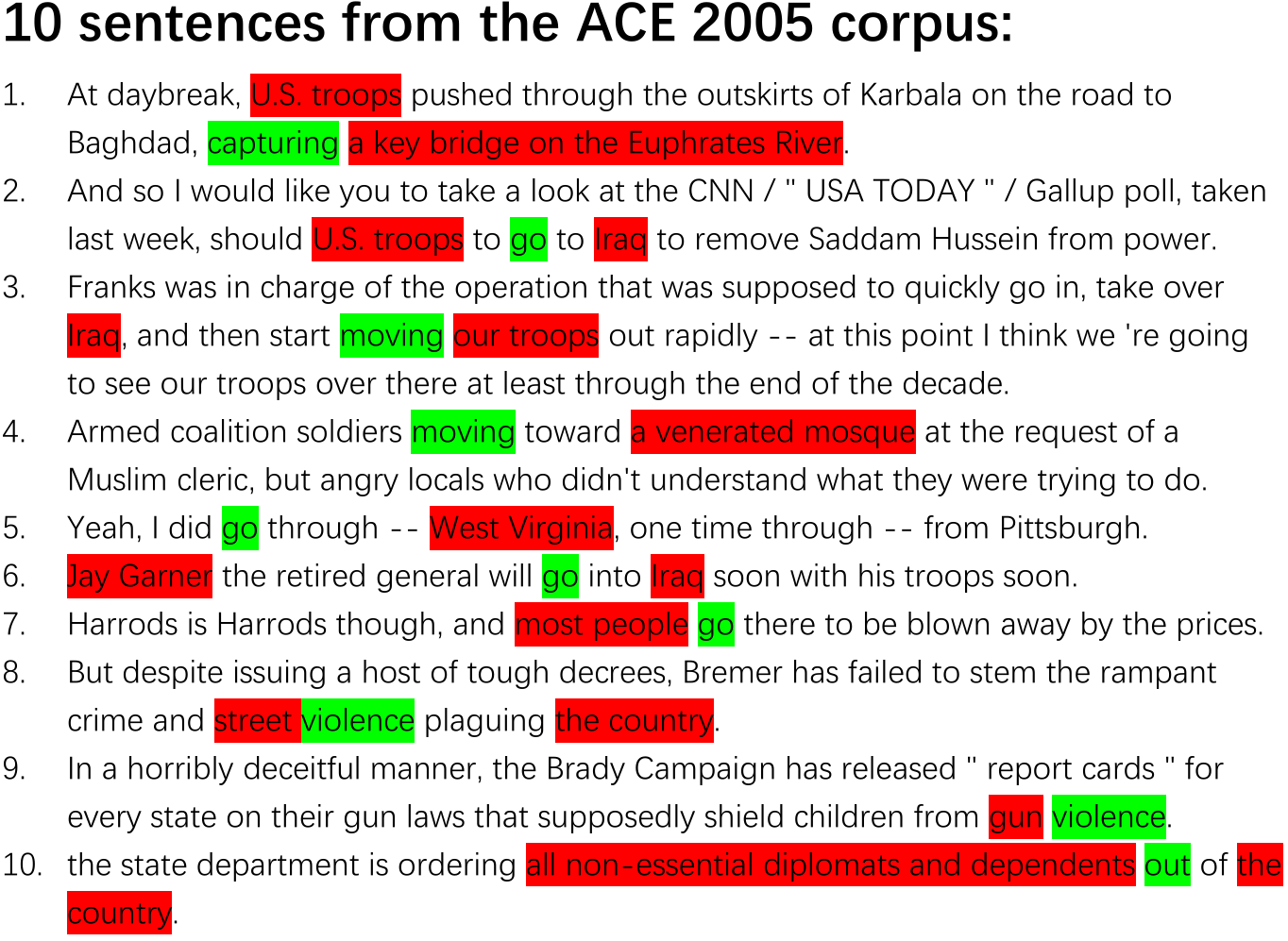}
	\caption{}
	\label{fig:illustrated_b} 
\end{subfigure}
\begin{subfigure}{2.4in}
		\includegraphics[width=\textwidth]{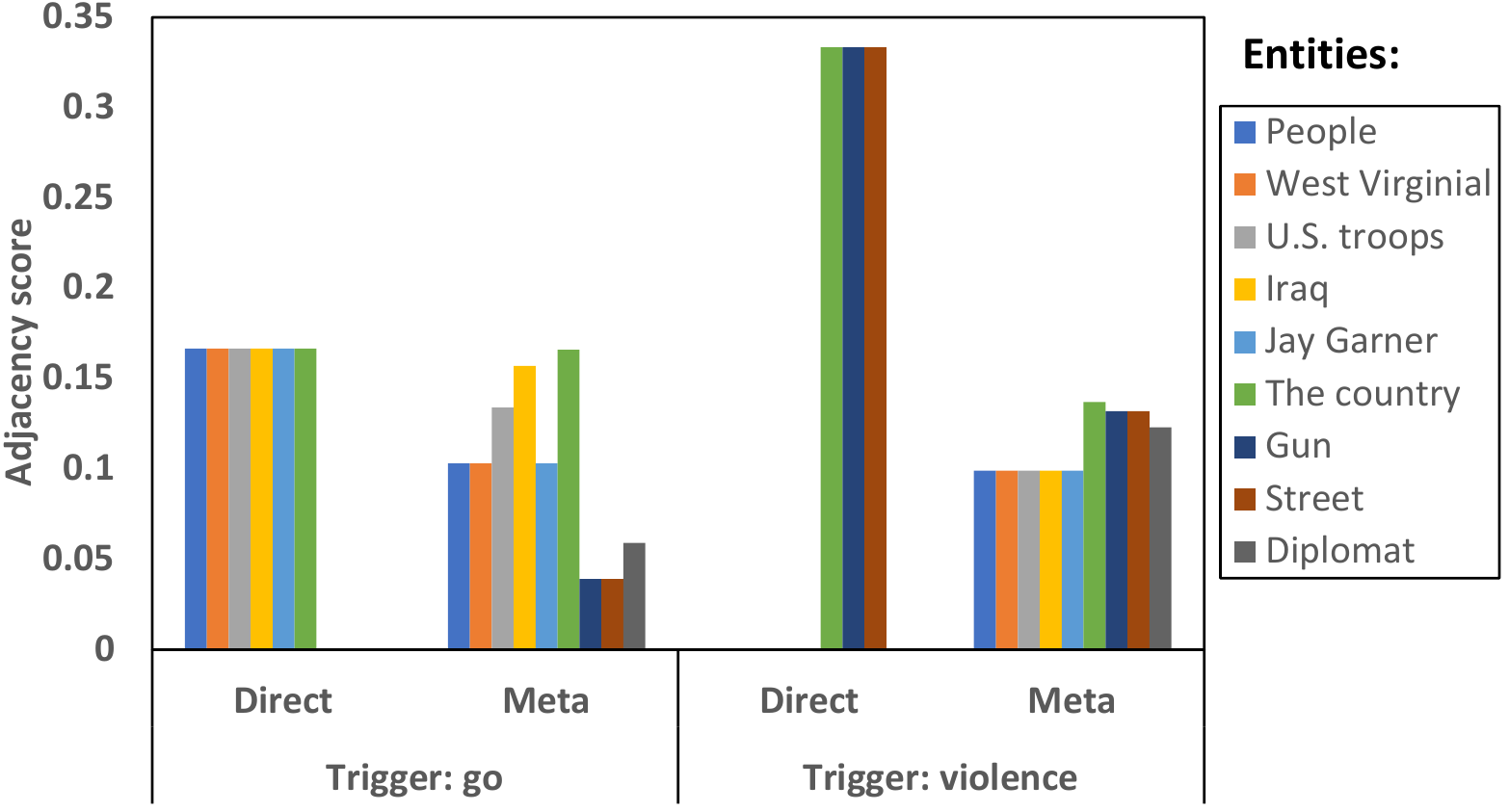}
	\caption{}
	\label{fig:illustrated_c} 
\end{subfigure}
\caption{An example of the meta-path based distribution on a heterogeneous information network (HIN). (a) The ``entity-trigger'' HIN for 10 sentences from the ACE 2005 corpus~\cite{ldc}, where green nodes are triggers and red nodes are entities; triggers are words that express the types of events in the texts (e.g. ``go'' and ``violence'' in this example). (b) The original 10 sentences for the HIN in this example. (c) Direct-adjacency-distribution for entities (Direct) v.s. meta-path-based distribution with path-length of 3 for entities (Meta) based on a given trigger. The meta-path-based distribution collects more indirect co-occurrence patterns for entities than the direct distribution (e.g. from ``go'' to ``gun'' along the meta-path ``Movement-GPE-Conflict-WEA''). The ``Movement'' and ``Conflict'' are predefined trigger types; ``GPE'', ``PER'' and ``WEA'' are predefined entity types for the geographical-social-political, person and weapon entities respectively. More information about entity and trigger types are referred to the document of the ACE 2005.}
\label{fig:illustrated_example} 
\end{figure*}

To address the aforementioned challenge, the core insight of this paper is that in the joint-event-extraction task, the ground-truth annotations for triggers could be leveraged to supervise the extraction of the entities, and vice versa. Based on this insight, this paper proposes a novel method to extract structural information from corpora by utilizing the co-occurrence relationships between triggers and entities. Furthermore, in order to
fully address the aforementioned sparsely co-occurrence relationships, we model the entity-trigger co-occurrence pairs as a heterogeneous information network (HIN) and supervise the trigger extraction by inferring the entity distribution with given triggers based on the indirect co-occurrence relationships collected along the meta-paths from a heterogeneous information network (HIN).

Figure~\ref{fig:illustrated_example} illustrates the process of our proposed method to collect indirect co-occurrence relationships between entities and triggers. Figure~\ref{fig:illustrated_a} is a sub-graph of the ``entity-trigger'' HIN for the ACE 2005 corpus~\cite{ldc}. Figure~\ref{fig:illustrated_c} compares the entity distributions inferred from given triggers based on the direct adjacency matrix and that inferred from the meta-path adjacency matrix. From this figure, we observe that a trigger does not necessarily connect to all entities directly and the direct-adjacency-based distribution is more concentrated on a few entities, while the meta-path-based distribution is spread over a larger number of entities. This shows that a model could collect indirect co-occurrence patterns between entities and triggers based on the meta-path adjacency matrix of an ``entity-trigger'' HIN. Moreover, the obtained indirect patterns could be applied to improve the performance to extract both entities and triggers.

Based on the aforementioned example and analysis, we propose a neural network to extract event entities and triggers. Our model is built on the top of sequence-to-sequence labeling framework and its inner parameters are supervised by both the ground-truth annotations of sentences and ``entity-trigger'' co-occurrence relationships. Furthermore, to fully address the indirect ``entity-trigger'' co-occurrence relationships, we propose the \underline{C}ross-\underline{S}upervised \underline{M}echanism (CSM) based on the HIN. The CSM alternatively supervises the entity and trigger extraction with the indirect co-occurrence patterns mined from a corpus. CSM builds a bridge for triggers or entities by collecting their latent co-occurrence patterns along meta-paths of the corresponding heterogeneous information network for a corpus. Then the obtained patterns are applied to boost the performances of entity and triggers extractions alternatively. We define this process as a ``cross-supervise'' mechanism. The experimental results show that our method achieves higher precisions and recalls than several state-of-the-art methods.

In summary, the main contributions of this paper are as follows:
\begin{itemize}
    \item We formalize the joint-event-extraction task as a sequence-to-sequence labeling with a combined tag-set, and then design a novel model, CSM, by considering the indirect ``entity-trigger'' co-occurrence relationships to improve the performance of joint-event-extraction.
    \item We are the first to use the indirect ``entity-trigger'' co-occurrence relationships (encoded in HIN) to improve the performance of the joint-event-extraction task. With the co-occurrence relationships collected based on meta-path technology, our model can be more precise than the current methods without any predefined features.
    \item Our experiments on real-world datasets show that, with the proposed cross-supervised mechanism, our method achieves better performance on the joint-event-extraction task than other related alternatives.
\end{itemize}

The remainder of this paper is organized as follows. In Section~\ref{sec_preliminaries}, we first introduce some preliminary knowledge about event extraction and HIN, and also formulate the problem. Section~\ref{sec_framework} presents our proposed model in detail. Section~\ref{sec_experiment} verifies the effectiveness of our model and compares it with state-of-the-art methods on real-world datasets. Finally, we conclude this paper in Section~\ref{sec_conclusion}.

\section{Preliminaries}\label{sec_preliminaries}
We formalize the related notations about the joint-event-extraction and heterogeneous information network.


\subsection{The Joint-Event-Extraction Task}

The sequence-to-sequence is a popular framework for event extraction \cite{DBLP:journals/tacl/ChiuN16}, which has been widely adopted in many recent related works. These methods annotate each token of a sentence as one tag in a pre-defined tag-set $\mathcal{A}$. In this way, a model based on sequence-to-sequence framework learns the relationship between original sentences and annotated tag-sequences.
Recurrent Neural Networks (RNN)~\cite{DBLP:conf/nips/SutskeverVL14} have shown promising performance in dealing with sequence-to-sequence learning problems. Therefore, lots of recent works~\cite{MaH16,NguyenCG16} apply RNN to perform the sequence-to-sequence event extraction.

\textit{Combined Annotation Tag-Set.}
In order to extract the entities and trigger words jointly under the sequence-to-sequence framework, one way is to extend the original tag-set $\mathcal{A}$ to a combined tag-set of entity types and trigger types, i.e. $\mathcal{A}=\mathcal{A}_e\bigcup{\mathcal{A}_t}$, where $\mathcal{A}_e$ and $\mathcal{A}_t$ represent the set of entity types and trigger types, respectively.

%
%
%

Given a sentence $s=\{w_1, w_2,\ldots,w_n\}$, where $w_i$s are tokens ($i=1,2,\ldots,n$), the joint-event-extraction is defined as the process to annotate each $w_i$ ($w_i\in{s}$) as one of the tags in set $\mathcal{A}$. This results in an annotated sequence $\phi(s)=\{y_1,y_2,\ldots,y_n\}$, where $y_i\in{\mathcal{A}}$. Then the joint event extraction becomes a sequence-to-sequence labeling~\cite{MaH16} which transforms a token sequence into a tag sequence.

\textit{Sequence-to-Sequence Labeling.} The goal of joint-event-extraction is to train a machine learning model under the supervision of a pre-annotated corpus. Minimizing the cross-entropy loss function~\cite{DBLP:journals/anor/BoerKMR05} has always been introduced to achieve this goal. The cross-entropy loss function is defined as follows:
\begin{equation}
\begin{split}
\mathcal{L}=\arg\min\sum_{\forall i\in{[1,n]}}\sum_{\forall{y_i\in{\mathcal{A}}}}-Pr(y_i|w_i)\log(\hat{Pr}(y_i|w_i)),
\end{split}
\label{seq2seqlabel}
\end{equation}
where $\hat{Pr}(y_i|w_i)$ is the probability for a model to annotate a token $w_i$ as a tag $y$ and $Pr(y_i|w_i)$ is the probability of an oracle model to annotate the token $w_i$ as the tag $y_i$ ($\forall{y_i}\in{\mathcal{A}}$). Within the framework of sequence-to-sequence labeling, entities and triggers could be recognized simultaneously by mapping the token sequence (of a sentence) to a combined tag sequence.

Generally, an event is modeled as a structure consisting of elements, such as event triggers and entities in different roles~\cite{NguyenCG16}. As shown in Figure~\ref{fig:illustrated_example}, event factors~\cite{DBLP:conf/acl/ChenLZLZ17} from sentences accumulate to a heterogeneous information network~\cite{DBLP:journals/tkde/ShiLZSY17} with nodes in different types. Furthermore, we observe that all edges or direct connections in Figure~\ref{fig:illustrated_example} are between triggers and entities, implying that named entities and triggers are contexts for each other. Intuitively, the performance of a joint-event-extraction task may degrade if it annotates triggers without the supervision of entities or annotates entities without the supervision of triggers.

\subsection{``Entity-Trigger'' Heterogeneous Information Network}

Given a corpus $\mathcal{D}$, an ``entity-trigger'' heterogeneous information network (HIN) is a weighted graph ${G}=\langle{V, {E}, {W}}\rangle$, where $V$ is a node set of entities and triggers; ${E}$ is an edge set, for $\forall{e_{i,j}}\in{E}$ ($e_{i,j}=\langle{v_i,v_j}\rangle$ , $v_i,v_j\in{V}$), $e_{i,j}$ denotes that $v_i$ and $v_j$ are co-occurred in a sentence of $D$; ${W}$ is a set of weight, for $\forall{w_{i,j}\in{W}}$, $w_{i,j}=w(v_i,v_j)$ ($v_i,v_j\in{V}$), $w_{i,j}$ refers to the frequency that $v_i$ and $v_j$ are co-occurred in sentences of $\mathcal{D}$. Furthermore, ${G}$ contains a node type mapping function $\phi:V\rightarrow{\mathcal{A}}$ and a link type mapping function $\psi:E\rightarrow{\mathcal{R}}$, where $\mathcal{A}$ is the combined annotation tag-set and $\mathcal{R}$ denotes the set of predefined ink types.

In particular, an ``entity-trigger'' HIN can be obtained by treating co-occurrence relationships between entities and triggers as edges. As illustrated in Figure~\ref{fig:illustrated_example}, ``entity-trigger'' HINs are usually sparse since entities do not directly connect (or co-occur) to all triggers and vice versa. In order to collect this indirect information, we resort to the meta-path~\cite{DBLP:journals/tkde/ShiLZSY17} based on ``entity-trigger'' HIN.

\textit{Meta-Path~\cite{DBLP:journals/tkde/ShiLZSY17}.} A meta-path is a sequence $\rho=\mathcal{A}_1\stackrel{\mathcal{R}_1}\longrightarrow{\mathcal{A}_2}\stackrel{\mathcal{R}_2}\longrightarrow\cdots\stackrel{\mathcal{R}_l}\longrightarrow{\mathcal{A}_{l+1}}$, where $l$ is the length of this path and $\mathcal{A}_i\in{\mathcal{A}}$ ($i=1,2,\ldots,l+1$). Generally, $\rho$ could be abbreviated as $\mathcal{A}_1\mathcal{A}_2\ldots{\mathcal{A}}_{l+1}$.

\begin{example}
As shown in Figure~\ref{fig:illustrated_a}, given two basic paths ``U.S. troops-go-Iraq'', ``most people-go-the country'' in the ACE 2005 corpus~\cite{ldc}, the corresponding meta-path is ``PER-Movement-GPE'' for both basic paths, where ``Movement'' is a trigger type, ``PER'' and ``GPE'' are entity types. This observation shows that the entities in types ``PER'' and ``GPE'' are indirectly connected through the given meta-path in the ACE 2005.
\end{example}

Since the routes for meta-paths are node types, they are much more general than direct paths. Furthermore, the meta-paths encode the indirect co-occurrence relationships between triggers and entities. Therefore, we can collect the latent information in the ``entity-trigger'' HIN along meta-paths to alleviate the sparse co-occurrence issue between entities and triggers.

\subsection{Problem Formulation}

In this section, we formalize the problem of joint-event-extraction by utilizing the co-occurrence relationships between entities and triggers (abbreviated as co-occurrence relationships in the following part) in a HIN.

\textit{Joint-Event-Extraction via HIN}. Given a corpus $\mathcal{D}$, its ``entity-trigger'' HIN ${G}$ and a set of meta-paths $\varrho$. The task of joint-event-extraction via HIN is to map the token sequences (of sentences) in $\mathcal{D}$ to sequences of tags (for any tag $\forall{y}\in\mathcal{A}$) with the co-occurrence patterns in ${G}$ based on the meta-paths in $\varrho$.

Intuitively, the corresponding ``entity-trigger'' HIN of a given corpus is naturally aligned together to form a knowledge graph that conforms to a corpus and can be used to supervise both the extracting processes for named entities and event triggers. In other words, if an annotation (both for entities and triggers) from a corpus violates its corresponding ``entity-trigger'' HIN, the entities and triggers in this result must be ill-annotated.

\section{Our Proposed Model}\label{sec_framework}
\begin{figure*}[htb]
	\centering
		\includegraphics[width=5.3in]{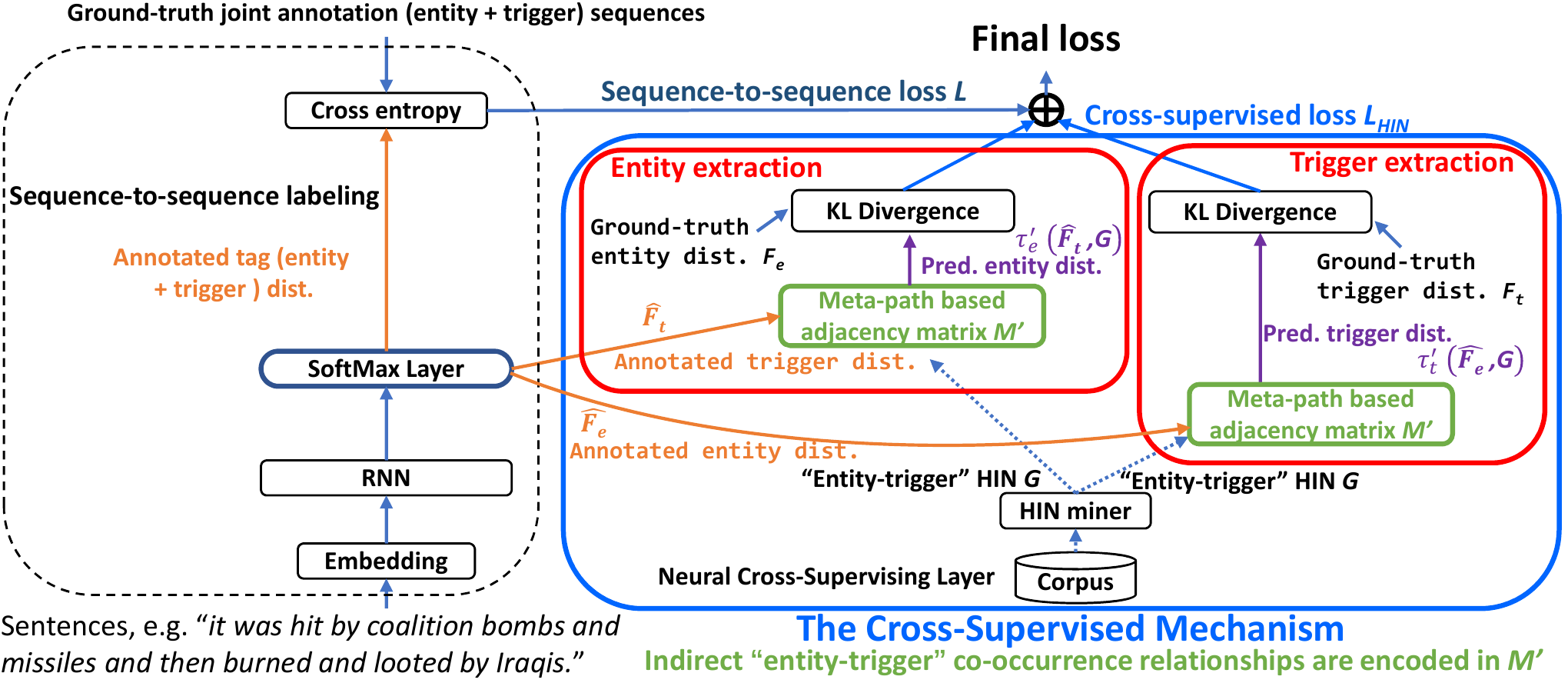}
	\caption{The framework of the joint-event-extraction model with our proposed cross-supervised mechanism.}
	\label{fig:framework} 
    \vspace{-0.1in}
\end{figure*}
As shown in Figure~\ref{fig:framework}, we define our task as a two-step process. First, it performs sequence-to-sequence labeling to annotate all entities and triggers, as shown on the left hand side of Figure~\ref{fig:framework}. Then, it supervises the annotated results by inferring the probabilities of the predicted entity and trigger based on the annotated results and indirect co-occurrence relationships, as shown on the right hand side of Figure~\ref{fig:illustrated_a}. To predict the entities or triggers distributions, we propose the meta-path based adjacency matrix for a given HIN and apply it to alternatively derive the entity and trigger distributions from each other. We name our method as the \underline{C}ross-\underline{S}upervised \underline{M}echanism (CSM) and implement it by a well designed neural cross-supervised layer (NCSL). Moreover, since the NCSL can be linked with any differentiable loss function, it can also be easily extended to many other event-extraction models. In this section, we will elaborate each part of our proposed model.

\subsection{Cross-Supervised Mechanism}



To incorporate the co-occurrence relationship into the joint-event-extraction process, we propose the cross-supervised mechanism. It is based on the observation that triggers and entities are prevalently connected in an ``entity-trigger'' HIN (cf. Figure~\ref{fig:illustrated_example}). With this observation, in a given corpus, the trigger of an event indicates the related entities. Meanwhile, the entities of an event also contain evidence for the corresponding trigger. Therefore, an extracted result could be evaluated by comparing the predicted entities (or triggers) based on the extracted triggers (or entities) with ground-truth entities (triggers). In order to implement this idea, we first define the probability distributions for entities and triggers.

\textbf{Entity and Trigger Distribution.} The entity distribution $F_e(x)=Pr(x=a)$ is a probability function for any entity type $\forall{a}\in{\mathcal{A}_e}$, while the trigger distribution $F_t(x)=Pr(x=a)$ is a probability function for any trigger type $\forall{a}\in{\mathcal{A}_t}$.
With these notations of entity and trigger distributions, the cross-supervised mechanism could be defined as follows.

\textit{Cross-Supervised Mechanism.} Given an entity distribution $F_e(x)$, a trigger distribution $F_t(x)$ for a corpus $\mathcal{D}$ and the corresponding HIN $G$; Suppose $\hat{F}_e(x)$ and $\hat{F}_t(x)$ are entity and trigger distributions based on the extraction results of a model. Then the target of cross-supervised mechanism is to minimize the following loss function:
\begin{equation}
\small
\mathcal{L}_{H\!I\!N}\!=\Delta(\tau_e(\hat{F}_t(x),G),F_e(x))\!+\Delta(\tau_t(\hat{F}_e(x),G),F_t(x)),
\label{diff}
\end{equation}
where $\tau_e(\hat{F}_t(x),G)$ and $\tau_t(\hat{F}_e(x),G)$ are the functions to predict entity and trigger distributions with the extracted results based on $G$; $\Delta$ is a function to compute the difference between two distributions. Intuitively, $\mathcal{L}_{H\!I\!N}$ measures the loss between the predicted and ground-truth distributions for entities and triggers.

To alternatively predict the entities (or triggers) based on the given triggers (or entities) from a HIN, the adjacency matrix of ``entity-trigger'' HIN is a natural tool to convert one (e.g. entity or trigger) distribution to another.

\textit{Entity-Trigger Direct Adjacency Matrix.} The entity-trigger direct adjacency matrix is an $\mathbb{R}^{\|\mathcal{A}_e\|\times\|\mathcal{A}_t\|}$ matrix $M=\{m_{i,j}\}_{\|\mathcal{A}_e\|\times\|\mathcal{A}_t\|}$, where $m_{u,v}=m_{u,v}$ refers to the frequency that an entity $u$ and a trigger $v$ are co-occurred in sentences of a corpus.

With the notation of the entity-trigger direct adjacency matrix, the alternative predicting function $\tau_t(\hat{F}_e,M)$ and $\tau_e(\hat{F}_t,M)$ can be computed as the following equations:
\begin{equation}
\tau_t(\hat{F}_e,G)=\hat{F}_e\times{M}^T,
\label{eq:convert1}
\end{equation}
\begin{equation}
\tau_e(\hat{F}_t,G)=\hat{F}_t\times{M},
\label{eq:convert2}
\end{equation}
where $F_e$ and $\hat{F}_e$ are $\mathbb{R}^{\|\mathcal{A}_e\|}$ vectors; $F_t$ and $\hat{F}_t$ are $\mathbb{R}^{\|\mathcal{A}_t\|}$ vectors; $F_e=[F_e(x_1),F_e(x_2),\ldots,F_e(x_{_{\|\mathcal{A}_e\|}})]$ and $\hat{F}_e=[\hat{F}_e(x_1),\hat{F}_e(x_2),\ldots,\hat{F}_e(x_{_{\|\mathcal{A}_e\|}})]$ for $\forall{x_i}\in{\mathcal{A}}_e$; $F_t=[F_t(x_1),F_t(x_2),\ldots,F_t(x_{_{\|\mathcal{A}_t\|}})]$ and $\hat{F}_t=[\hat{F}_t(x_1),\hat{F}_t(x_2),\ldots,\hat{F}_t(x_{_{\|\mathcal{A}_t\|}})]$  for $\forall{x_i}\in{\mathcal{A}}_t$.
However, since the ``entity-trigger'' HIN may be sparse (cf. Figure \ref{fig:illustrated_c}),
it is challenging to precisely predict entity and trigger distributions with inadequate evidence. Thus, we resort to the meta-path based technology to utilize the sparse information in a HIN.

\textit{Meta-Path based Adjacency Matrix.} In the same setting of the direct adjacency matrix, given a set of meta-paths $\varrho$, the meta-path based adjacency matrix is an $\mathbb{R}^{\|\mathcal{A}_e\|\times\|\mathcal{A}_t\|}$ matrix $M'=\{m'_{u,v}\}_{\|\mathcal{A}_e\|\times\|\mathcal{A}_t\|}$, where $m'_{u,v}$ is denoted as:
\begin{equation}
m'_{u,v}=\sum_{\rho\in\varrho}{\log{Pr_{\rho}(u,v)}},
\end{equation}
where $Pr_{\rho}(u,v)$ is the reachable probability from $u$ to $v$ based on a given meta-path $\rho$. Suppose $\|\rho\|=l$, $Pr_{\rho}({u,v})$ is computed in the following equation:
\begin{equation}
Pr_{\rho}({u,v})=\prod^{n_1=u,n_l=v}_{i\in[1,l],\phi(n_i)=\rho_i}w_{n_{i+1,i}}{Pr(n_{i+1}|n_{i})},
\end{equation}
where $\phi(n_i)$ is the type of node $n_i$, $\rho_i$ is the $i$-th type in path $\rho$ ($\rho_{i}\in{\mathcal{A}_e}$); $w_{n_{{i+1},i}}$ is the frequency that $n_i$ and $n_{i+1}$ are co-occurred in sentences; $Pr(n_{i+1}|n_{i})$ is the reachable probability from node $n_i$ to $n_{i+1}$ by considering the types $\phi{(n_i)}$ and $\phi{(n_{i+1})}$. $Pr(n_{i+1}|n_{i})$ can be obtained through a meta-path based random walk~\cite{DBLP:journals/tkde/ShiHZY19}.

\begin{equation}
Pr(n_{i+1}|n_{i})=
\begin{cases}
\frac{1}{|N^{\rho_{i+1}}(n_i)|}, |N^{\rho_{i+1}}(n_i)|>0\\
0,else,
\end{cases}
\end{equation}
where $N^{\rho_{i+1}}(n_i)$ is the set of direct neighbors for node $n_i$ by considering the next type $\rho_{i+1}$ on path $\rho$.

By replacing the adjacency matrices as meta-path based adjacency matrices in Eq.~\ref{eq:convert1} and Eq.~\ref{eq:convert2}, the entity and trigger distributions can be predicted through the following equations:

\begin{equation}
\tau_t'(\hat{F}_e,G)=\hat{F}_e\times{M'}^T,
\label{eq:convert1p}
\end{equation}

\begin{equation}
\tau_e'(\hat{F}_t,G)=\hat{F}_t\times{M'},
\label{eq:convert2p}
\end{equation}
where $\tau_t'(\hat{F}_e,G)$ and $\tau_e'(\hat{F}_t,G)$ compute the entity and trigger meta-path based distributions, respectively.

\subsection{Neural Cross-Supervised Layer}

With the aforementioned discussion, we could further evaluate the possibility of the trigger distribution based on the annotated entities of a model or evaluate the possibility that the entity distribution of the entity distribution based on the annotated triggers of the same model. We name this evaluation process as the cross-supervision and implement it in the NCSL. By substituting the Eq.~\ref{eq:convert1p} and Eq.~\ref{eq:convert2p} for corresponding terms in Eq.~\ref{diff}, NCSL evaluates this difference with two concatenate KL-divergence loss~\cite{DBLP:conf/iccv/GoldbergerGG03} in the following:
\begin{equation}
\mathcal{L}_{H\!I\!N}\!=\!{F_t\log{\frac{F_t}{\tau_t'(\hat{F}_e,G)}}}+\!{F_e\log{\frac{F_e}{\tau_e'(\hat{F}_t,G)}}},
\label{cross}
\end{equation}
where $\hat{F}_e$ and $\hat{F}_t$ are the predicted distributions for entities and triggers by the sequence-to-sequence labeling; $F_e$ and $F_t$ are the ground-truth entity and trigger distributions, respectively.
In this way, NCSL incorporates both the cross-supervised information for triggers and entities into its process.

\subsection{Training the Complete Model}

We formalize the complete process of our model as follows.

\textit{Cross-Supervised Joint-event-extraction.} The objective of our task is to optimize the following equation:
\begin{equation}
    \mathcal{L}_c=(1-\alpha){\mathcal{L}} + \alpha{\mathcal{L}_{H\!I\!N}},
    \label{eq:loss_complete}
\end{equation}

where $\mathcal{L}$ is the loss for a sequence-to-sequence labeling in Eq.~\ref{seq2seqlabel}, $\mathcal{L}_{H\!I\!N}$ is the loss for the cross-supervised process in Eq.~\ref{cross} and $\alpha$ is the ratio for the cross-supervised process.

As illustrated in Figure~\ref{fig:framework}, this model implements the sequence-to-sequence labeling with an embedding layer which embeds the input sentences as sequences of vectors and a Bidirectional Long-Short-Term Memory (Bi-LSTM) network~\cite{DBLP:conf/aclnut/LimsopathamC16} of RNN~\cite{DBLP:conf/nips/SutskeverVL14} family to predict the tag distribution based on the embedded vector sequences. The training applies the back-propagation with the Adam optimizer~\cite{DBLP:journals/corr/KingmaB14} to optimize this loss function.

\subsection{Discussion}\label{sec_framework_novelty}

From Eq. \ref{eq:loss_complete}, we observe that our task is equivalent to the sequence-to-sequence method when ${\alpha}=0$. Therefore, our model could be easily implemented by following an end-to-end framework with extra supervision information incorporated in the co-occurrence relationships.
Here we also summarize the novelty of our proposed approach as the introduced cross-supervised mechanism by incorporating indirect co-occurrence relationships collected from the ``entity-trigger'' HIN along meta-paths (cf. $\mathcal{L}_{HIN}$ in Eq. \ref{eq:loss_complete}), for the task of joint-event-extraction. The introduced cross-supervised mechanism aims to maximizing the utilization efficiency of the training data, so that more effective information will be considered to improve the performance of joint-event-extraction.

\begin{table}[h]
	\centering
	\caption{Dataset statistics}
\begin{tabular}{ccccc}
	\toprule
		 &ACE2005&NYT&CoNLL&WebNLG\\
	\midrule
sentences&2,107&6,304&3,932&10,165\\
entities&4,590&12,643&13,511&2,217\\
triggers&1,921&6,355&3,903&1,309\\
entity types&11&17&4&9\\
event types&8&4&11&289\\
meta-paths (l=3)&4,459&18,035&22,399&12,675\\
	\bottomrule
	\end{tabular}
	\label{table_data_statistic}
\end{table}
\section{Experiment and Analysis}\label{sec_experiment}
We compare our model with some state-of-the-art methods to verify the effectiveness of the proposed mechanism.
\begin{table*}[t]
	\centering
		\caption{Comparison on real-world datasets}
\scalebox{0.7}[0.7]{
        \begin{tabular}{ccccccccccccc}
        	\toprule
            \multirow{2}*{Model}&\multicolumn{3}{c}{ACE 2005}&\multicolumn{3}{c}{NYT}&\multicolumn{3}{c}{CoNLL}&\multicolumn{3}{c}{WebNLG}\\
            \cline{2-13}~&Precision&Recall&F1&Precision&Recall&F1&Precision&Recall&F1&Precision&Recall&F1\\
            \midrule
Seq2Seq&0.442$\pm$0.025&0.493$\pm$0.0272&0.466$\pm$0.026&0.818$\pm$0.012&0.832$\pm$0.012&0.825$\pm$0.012&0.709$\pm$0.015&0.852$\pm$0.011&0.774$\pm$0.013&0.851$\pm$0.009&0.910$\pm$0.007&0.880$\pm$0.008\\
CRF&0.434$\pm$0.031&0.478$\pm$0.033&0.455$\pm$0.032&0.813$\pm$0.011&0.828$\pm$0.011&0.821$\pm$0.01&0.718$\pm$0.016&0.867$\pm$0.013&0.785$\pm$0.014&0.864$\pm$0.005&0.921$\pm$0.005&0.892$\pm$0.005\\
GCN&0.435$\pm$0.030&0.487$\pm$0.032&0.459$\pm$0.031&0.804$\pm$0.013&0.819$\pm$0.013&0.811$\pm$0.013&0.706$\pm$0.015&0.871$\pm$0.014&0.780$\pm$0.013&0.884$\pm$0.008&0.931$\pm$0.008&0.907$\pm$0.008\\
JEE&0.423$\pm$0.023&0.468$\pm$0.030&0.443$\pm$0.026&0.717$\pm$0.009&0.645$\pm$0.014&0.679$\pm$0.012&0.713$\pm$0.019&0.814$\pm$0.013&0.76$\pm$0.015&0.775$\pm$0.015&0.818$\pm$0.012&0.796$\pm$0.013\\
JT&0.469$\pm$0.003&0.426$\pm$0.005&0.447$\pm$0.004&0.725$\pm$0.012&0.691$\pm$0.006&0.708$\pm$0.009&0.738$\pm$0.025&0.837$\pm$0.006&0.784$\pm$0.021&0.818$\pm$0.011&0.829$\pm$0.007&0.823$\pm$0.008\\
CSM$\rm _{DA}$&0.455$\pm$0.024&0.494$\pm$0.022&0.474$\pm$0.023&0.835$\pm$0.012&0.847$\pm$0.012&0.841$\pm$0.012&0.730$\pm$0.017&0.856$\pm$0.021&0.788$\pm$0.019&0.908$\pm$0.005&0.941$\pm$0.004&0.924$\pm$0.004\\
CSM$\rm_{HIN}$&\textcolor{blue}{0.477$\pm$0.030}&\textcolor{blue}{0.533$\pm$0.033}&\textcolor{blue}{0.503$\pm$0.031}&\textcolor{blue}{0.859$\pm$0.007}&\textcolor{blue}{0.870$\pm$0.008}&\textcolor{blue}{0.865$\pm$0.008}&\textcolor{blue}{0.754$\pm$0.018}&\textcolor{blue}{0.890$\pm$0.020}&\textcolor{blue}{0.816$\pm$0.017}&\textcolor{blue}{0.923$\pm$0.004}&\textcolor{blue}{0.953$\pm$0.003}&\textcolor{blue}{0.937$\pm$0.003}\\
        	\bottomrule
        \end{tabular}
        }
	\label{table_comparison}
\end{table*}

\begin{table}[t]
	\centering
	\caption{Detailed comparison on ACE 2005}
\scalebox{0.9}[0.9]{
        \begin{tabular}{ccccccc}
        	\toprule
            \multirow{2}*{Model}&\multicolumn{3}{c}{Entity extraction}&\multicolumn{3}{c}{Trigger extraction}\\
            \cline{2-7}~&Precision&Recall&F1&Precision&Recall&F1\\
            \midrule
Seq2Seq&0.494&0.489&0.49&0.383&0.426&0.403\\
CRF&0.502&0.483&0.491&0.395&0.473&0.431\\
GCN&0.508&0.491&0.499&0.381&0.443&0.410\\
JEE&0.451&0.497&0.472&0.407&0.411&0.409\\
JT&0.492&0.458&0.474&0.447&0.414&0.432\\
CSM$\rm_{DA}$&0.509&0.535&0.52&0.404&0.442&0.422\\
CSM$\rm_{HIN}$&{\color{blue}0.512}&{\color{blue}0.552}&{\color{blue}0.532}&{\color{blue}0.464}&{\color{blue}0.484}&{\color{blue}0.474}\\
        	\bottomrule
        \end{tabular}
        }
	\label{table_comparison_detail}
\end{table}

\subsection{Datasets}\label{sec_experiment_dataset}
We adopt four real-world datasets which are widely used to evaluate our model. ACE 2005 is a corpus developed by Linguistic Data Consortium (LDC)~\cite{ldc}. NYT is an annotated corpus provided by the New York Times Newsroom~\cite{nyt}. CoNLL 2002~\cite{conll} is a Spanish corpus made available by the Spanish EFE News Agency. WebNLG is a corpus introduced by Claire et al.~\cite{DBLP:conf/acl/GardentSNP17} in the challenge of natural language generation, which also consists the entity label. Note that all aforementioned datasets except ACE 2005 do not provide the original ground-truth trigger annotations. In the testing phase, since it requires ground-truth trigger annotations to measure the performances of models, we instead use CoreNLP\footnote{https://stanfordnlp.github.io/CoreNLP/} to create the corresponding trigger annotations for these datasets. More details of our datasets are shown in Table~\ref{table_data_statistic}.

\subsection{Comparison Baselines}\label{sec_experiment_baseline}

We compare our method with some state-of-the-art baselines for event extraction.

\begin{itemize}
\item \textit{Sequence-to-Sequence Joint Extraction (Seq2Seq)} \cite{DBLP:conf/aclnut/LimsopathamC16} \cite{DBLP:conf/acl/ZhengWBHZX17} is a joint extraction method implemented by us in the sequence-to-sequence framework with a joint tag set contains tags for both entities and triggers.
\item \textit{Conditional Random Field Joint Extraction (CRF)} \cite{DBLP:conf/www/AlzaidyCG19} extends from the basic sequence-to-sequence framework with a conditional random field (CRF) layer which constraints the output tag orders.
\item \textit{GCN}~\cite{DBLP:conf/acl/FuLM19} jointly extracts entities and triggers by considering the context information with graph convolution network (GCN) layers behind the BiLSTM module.
\item \textit{Joint Event Extraction (JEE)} \cite{DBLP:conf/naacl/YangM16} is a joint statistical method based on the structural dependencies between entities and triggers.
\item \textit{Joint Transition (JT)}~\cite{DBLP:conf/ijcai/ZhangQZLJ19} models the parsing process for a sentence as a transition system, and proposes a neural transition framework to predict the future transition with the given tokens and learned transition system.
\item \textit{CSM$\rm _{DA}$} is the proposed model with Eq.~\ref{eq:convert1} and Eq.~\ref{eq:convert2} without considering the meta-paths.
\item \textit{CSM$\rm _{HIN}$} is our complete model with Eq.~\ref{eq:convert1p} and Eq.~\ref{eq:convert2p}.
\end{itemize}

\subsection{Evaluation Metrics}\label{sec_experiment_metrics}
To evaluate the performance of our proposed model, we adopt several prevalent metrics, e.g., precision, recall and F1 score, which have been widely used in the field of event extraction. The Precision and Recall are defined as follows:
\begin{equation}
    Precision = \frac{TP}{TP+FP},
\end{equation}
\begin{equation}
    Recall =  \frac{TP}{TP+FN},
\end{equation}
where $TP$ is the true positive frequency, $FP$ is the false positive frequency and $FN$ is the false negative frequency. The quantities $TP$, $FP$, and $FN$ are measured from the predicted tags of a model by referring to the ground-truth tags for the testing samples. In our setting, for a specific model, $TP$ records the number of predicted tags matching with the corresponding ground-truth tags for entities and triggers. $FP$, on the other hand, records the frequency of its predicted tags conflicting with the corresponding ground-truth tags, and $FN$ records the number of entities and triggers missed by a model.
\begin{equation}
    F1 = \frac{2\cdot{Precision\cdot{Recall}}}{Precision+Recall}.
\end{equation}
F1 measures the joint performance for a model by considering the precision and recall simultaneously.

\subsection{Implementation Details}\label{sec_experiment_implementation}

Since our aim is to incorporate the indirect co-occurrence relationships between the entities and their triggers into the joint-event-extraction task, not to investigate the influence of pre-trained features on different models, we implement all models in \ref{sec_experiment_baseline} without any pre-trained features on our prototype system. Furthermore, in order to compare all methods fairly, all the neural network models share the same LSTM module (a Bi-LSTM with 128 hidden dimensions and 2 hidden layers) as the basic semantic embedding. Moreover, all neural network models are trained through the Adam optimizer~\cite{DBLP:journals/corr/KingmaB14} with the same learning rate (0.02) and 30 training epoches. During the training, we set the embedding dimension of a word to 300, the batch size to 256, and the dropout to 0.5.

\textit{HIN Generation.} Our model requires HINs to convert between the entity and trigger distributions. We need to generate the required HINs in a preprocessing step. The HINs are generated by merging all ground-truth triggers and entities with their relationships and types from the training data. For each training process, the HIN is re-generated with different training data. During the testing process, the entity distribution is translated into the trigger distribution according to the corresponding HIN, without knowing any co-occurrence relationships between the entities and triggers in testing data. Moreover, our HINs are generated based on the basic event types since the obtained HINs based on event subtypes are too sparse to reveal effective indirect co-occurrence relationships.

In the following experiments, we compare the precision, recall and F1 scores for all methods in 10-fold cross-validation. The 10-fold cross-validation means we split the original data into 10 subsets randomly without intersection and train the models with 9 of these subsets. We test the models with the remaining subset. This procedure is repeated 10 times. We report the means and variances of the results in the remaining part. Furthermore, to compare the models on recognizing the effect event factors, we exclude the results for those tokens being labelled as the outside tag (or ``$O$'') for all methods.

\subsection{Experimental Results}\label{sec_experiment_result}

The results of the comparison experiment on all datasets are reported in Table~\ref{table_comparison}. We observe that with the cross-supervised mechanism provided by the NCSL layer, both CSM$\rm _{DA}$ and CSM$\rm _{HIN}$ surpass all the state-of-the-art methods.
Furthermore, we also measure the mean performances on entity and trigger extraction respectively using the ACE 2005 dataset for all methods. This result is reported in Table~\ref{table_comparison_detail}. We observe that our model outperforms the alternative models on both joint task and sub-tasks. This verifies that the extraction performance is indeed improved by the indirect co-occurrence relationships collected along the meta-paths of heterogeneous information networks.


\begin{figure}[htb]
	\centering
		\begin{subfigure}{1.4in}
		\includegraphics[width=\textwidth]{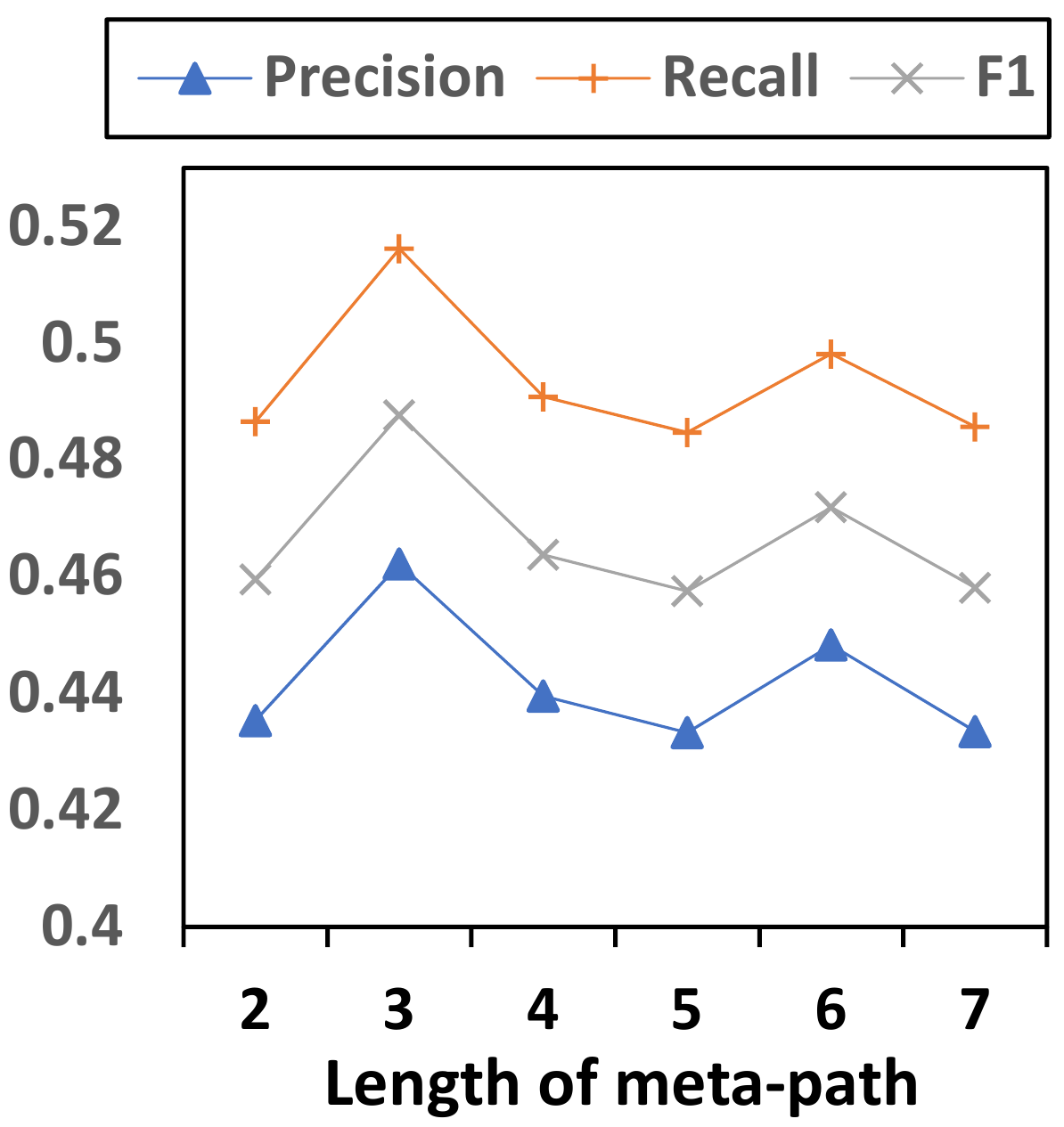}
	\caption{}
	\label{fig:sensitivity1} 
\end{subfigure}
		\begin{subfigure}{1.4in}
		\includegraphics[width=\textwidth]{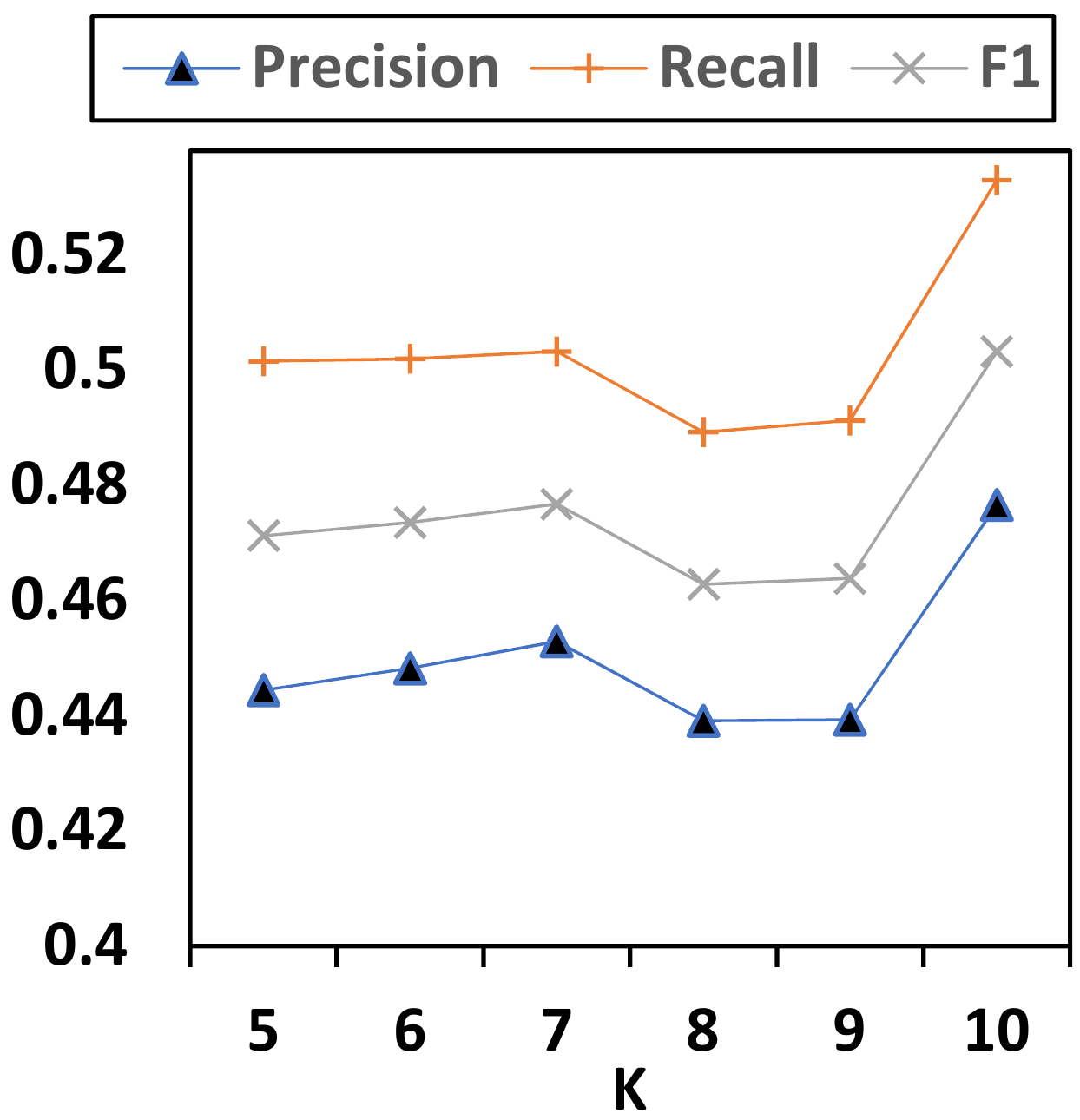}
	\caption{}
	\label{fig:sensitivity2} 
\end{subfigure}
	\caption{Sensitivity in different parameters}
	\label{fig:sensitivity}
\end{figure}
\vspace{-0.1in}

\subsection{Sensitivity Analysis}\label{sec_experiment_sensitivity}
We analyze the influence of the training ratio (from 5 to 10 fold cross-validation) and the length of meta-paths on the performance of our model. These experiments are performed on the
 ACE 2005 dataset and all of them are repeated 10 times. The mean results are reported in Figure~\ref{fig:sensitivity}. As shown in Figure~\ref{fig:sensitivity1}, our model achieves the best performance with the meta-path length of 3. The reason is that most of the ACE 2005 data are in the ``entity-trigger-entity'' form, our model performs well with the meta-path lengths which are multipliers of 3. Furthermore, from Figure \ref{fig:sensitivity2}, we can see our model also performs well when the $K$ is large, which confirms to the intuition that more training data lead to better performance.

\subsection{Case Study}\label{sec_experiment_casestudy}

\begin{figure}[htb]
	\centering
		\includegraphics[width=3.4in]{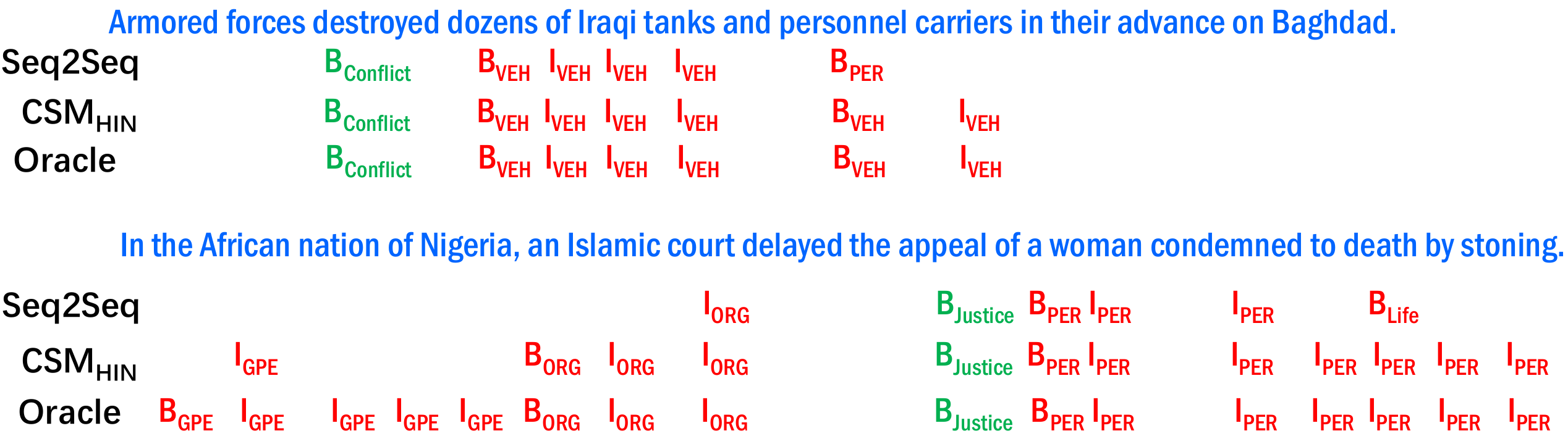}
	\caption{Part of annotation results on the ACE 2005 dataset.}
	\label{fig:casestudy}
\end{figure}

To figure out the improvement of our model on the extraction task, we focus on typical cases from the ACE 2005 dataset. These cases are presented in Figure~\ref{fig:casestudy}, where ``Oracle'' means the ground-truth annotation. We observe that in simple sentences, both the sequence-to-sequence method and our model annotate accurately. However, with the sentence becoming more complex (cf. the bottom sentence in Figure~\ref{fig:sensitivity}), the sequence-to-sequence method hardly annotates accurate entities that are far from the trigger, while our method keeps stable performance. This further shows that our method can extract the useful latent patterns along the meta-paths.

\vspace{-0.1in}

\section{Conclusion}\label{sec_conclusion}

In this paper, we have proposed a novel cross-supervised mechanism which allows models to extract entities and triggers jointly. Our mechanism alternately supervises the extraction process for either the triggers or the entities, based on the information in the type distribution of each other. In this way, we incorporate the co-occurrence relationships between entities and triggers into the joint-event-extraction process of our model. Moreover, to further address the problem caused by the sparse co-occurrence relationships, our method also resorts to the heterogeneous information network technology to collect indirect co-occurrence relationships. The empirical results show that our method improves the extraction performances for entities and triggers simultaneously. This verifies that the incorporated co-occurrence relationships are useful for the joint-event-extraction task and our method is more effective than existing methods in utilizing training samples. Our future works include: (a) investigating the impact of length of sampled meta-paths, as in this paper we have limited the meta-path into a fixed length; (b) connecting the extracted entities and triggers from a corpus to facilitate the automatic knowledge graph construction.

\section{Acknowledgements}
This work is supported by the National Natural Science Foundation of China (Grant No. 61976235, 61602535, 61503422), Program for Innovation Research in Central University of Finance and
Economics, and the Foundation of State Key Laboratory of Cognitive Intelligence (Grant No. COGOSC-20190002), iFLYTEK, P. R. China. This work is also supported in part by NSF under grants III-1763325, III-1909323, and SaTC-1930941.

\bibliographystyle{IEEEtran}
\bibliography{references}


%

\end{document}